\documentclass{article}

\usepackage{microtype}
\usepackage{graphicx}
\usepackage{subfigure}
\usepackage{booktabs} 
\usepackage{tikz}
\usetikzlibrary{positioning, arrows.meta, calc}
\usepackage{amsmath}
\usepackage{float}
\usepackage{wrapfig}
\usepackage{enumitem}
\usepackage{booktabs}
\usepackage{bm}

\usepackage{hyperref}

\usepackage[accepted]{icml2024}

\usepackage{amsmath}
\usepackage{amssymb}
\usepackage{mathtools}
\usepackage{amsthm}

\usepackage[capitalize,noabbrev]{cleveref}

\theoremstyle{plain}

\theoremstyle{definition}

\theoremstyle{remark}

\usepackage[textsize=tiny]{todonotes}

\icmltitlerunning{On Evaluation of Vision Datasets and Models using Human Competency Frameworks}

\begin{document}

\twocolumn[
\icmltitle{On Evaluation of Vision Datasets and Models using Human Competency Frameworks}



\icmlsetsymbol{equal}{*}

\begin{icmlauthorlist}
\icmlauthor{Rahul Ramachandran}{iith}
\icmlauthor{Tejal Kulkarni}{iith}
\icmlauthor{Charchit Sharma}{iith}
\icmlauthor{Deepak Vijaykeerthy}{ibm}
\icmlauthor{Vineeth N Balasubramanian}{iith}
\end{icmlauthorlist}

\icmlaffiliation{iith}{Indian Institute of Technology Hyderabad, India}
\icmlaffiliation{ibm}{IBM Research AI, India}

\icmlcorrespondingauthor{Rahul Ramachandran}{cs21btech11049@iith.ac.in}

\icmlkeywords{Machine Learning, ICML}

\vskip 0.3in
]



\printAffiliationsAndNotice{}  

\begin{abstract}
Evaluating models and datasets in computer vision remains a challenging task, with most leaderboards relying solely on accuracy. While accuracy is a popular metric for model evaluation, it provides only a coarse assessment by considering a single model's score on all dataset items. This paper explores Item Response Theory (IRT), a framework that infers interpretable latent parameters for an ensemble of models and each dataset item, enabling richer evaluation and analysis beyond the single accuracy number. Leveraging IRT, we assess model calibration, select informative data subsets, and demonstrate the usefulness of its latent parameters for analyzing and comparing models and datasets in computer vision.
\end{abstract}


\vspace{-9pt}
\section{Introduction}
\vspace{-2pt}
The fundamental goal of constructing datasets in computer vision is to accurately represent the true underlying data distribution, ensuring that good model performance translates into the ability to perform well on real-world tasks. Despite the progress facilitated by leaderboards, which rank models based on performance metrics like accuracy, this focus on state-of-the-art (SoTA) performance often obscures the true objective of improving overall model quality. Consequently, evaluating the ability of models and the quality of datasets remains a significant challenge.

Item Response Theory (IRT) \citep{baker_item:04}, a statistical framework traditionally used in educational assessment, has recently been adopted by the machine learning community to address these evaluation challenges. IRT models students' abilities and the difficulties of questions using latent parameters, providing nuanced insights into performance. Recent studies have begun leveraging IRT to gain a deeper understanding of datasets and models \citep{lalor_dynamic:20, vania_comparing:21, rodriguez_evaluation:21}.
While prior work has explored using IRT parameters for more nuanced leaderboards \citep{rodriguez_evaluation:21} and for analyzing dataset saturation, in this study, we conduct a comprehensive investigation that leverages IRT to provide insights into computer vision datasets like ImageNet, among others. We study model calibration by analyzing confidences through the IRT lens, using IRT parameters to analyze dataset quality and enable data-subset selection.

Our key contributions include: (i) We use 91 computer vision models and ImageNet dataset to extract latent IRT parameters such as Ability, Difficulty, Discriminability, and Guessing Parameters to provide insights into models and datasets (Sec \ref{sec_expts}); (ii) We define a new metric called overconfidence and demonstrate that strong models are well-calibrated; deviations from zero in overconfidence correlate with increased label errors, aiding in automatic annotation error detection. (Sec \ref{sec: model_calibration}); (iii) We explore the role of latent parameters in assessing dataset quality and complexity using the guessing parameter (Sec \ref{sec: dataset_complexity}); (iv) We leverage IRT to show that a sample of just 10 images can be used to discriminate between the relative performance of 91 models with a Kendall correlation of 0.85 (Sec \ref{sec: data_selection}).

\vspace{-7pt}
\section{Item Response Theory}

\vspace{2pt}
\noindent \textbf{Item Response Theory Models.}
The main objective of Item Response Theory (IRT) is to model the probability of an individual correctly responding to a given item or question. In our context, we leverage IRT to characterize the probability of a model correctly classifying an image. Specifically, the probability of model $i$ correctly classifying image $j$ can be described using three different IRT models, which are presented below:

    \begin{equation}\label{eq:1pl}
        p(y_{ij} = 1|\theta_i, b_j) = \frac{1}{1+ e^{-(\theta_i - b_j)}}
    \end{equation}
    
    \begin{equation}\label{eq:2pl}
        p(y_{ij} = 1|\theta_i, b_j, \gamma_j) = \frac{1}{1+ e^{-\gamma_j(\theta_i - b_j)}}
    \end{equation}

    \begin{equation}\label{eq:3pl}
        p(y_{ij} = 1|\theta_i, b_j, \gamma_j, \lambda_j) = \lambda_j + \frac{1 - \lambda_j}{1+ e^{-(\theta_i - b_j)}}
    \end{equation}

\begin{figure}[h]
    \centering
    \includegraphics[width=0.4\textwidth]{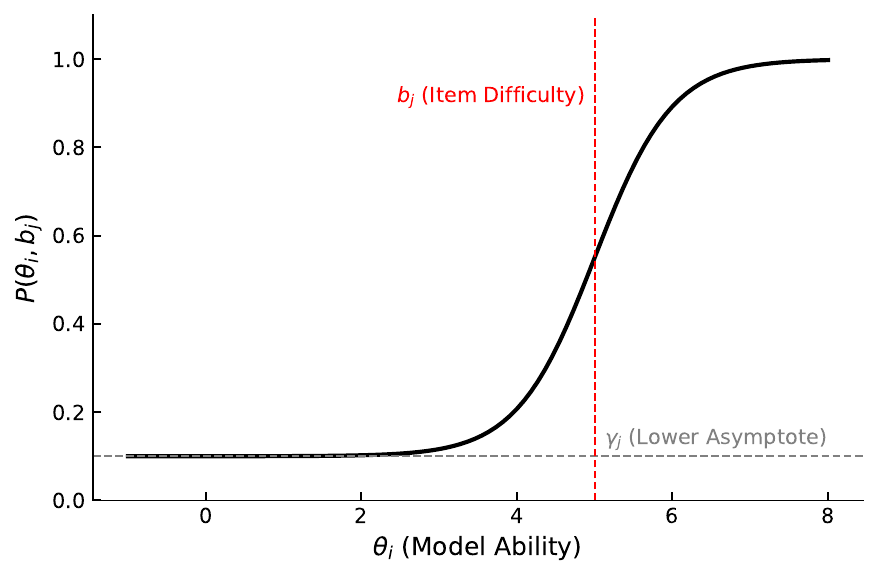}
    \vspace{-9pt}
    \caption{3PL ICC for image with $b=5$}
    \vspace{-6pt}
    \label{fig:icc}
\end{figure}

These models are termed the 1PL, 2PL and 3PL models respectively \citep{hambleton_item:85, baker_item:04}. The latent parameters $\theta, b, \gamma$ and $\lambda$ are called the \textbf{ability}, \textbf{difficulty}, \textbf{discriminability} and \textbf{guessing} parameters respectively. For a given image, we can plot the probabilities for each ability, giving us a curve known as the item characteristic curve (ICC). This curve is depicted in Fig. \ref{fig:icc}.
\vspace{-8pt}
As expected, the probability of classifying the image correctly increases monotonically with the model ability $\theta$. The difficulty $b$ determines the location of the curve, and the guessing parameter $\lambda$ determines the lower asymptote. The discriminability parameter $\gamma$ determines the steepness of the curve; an item with a high $\gamma$ distinguishes between models above and below the item
difficulty with a high probability. To infer the latent parameters, we employ variational inference. The overall workflow of our implementation is given below. For more details, refer to \ref{app:vi}.
\vspace{-8pt}
\begin{figure}[h]
    \centering
    \includegraphics[width=0.5\textwidth]{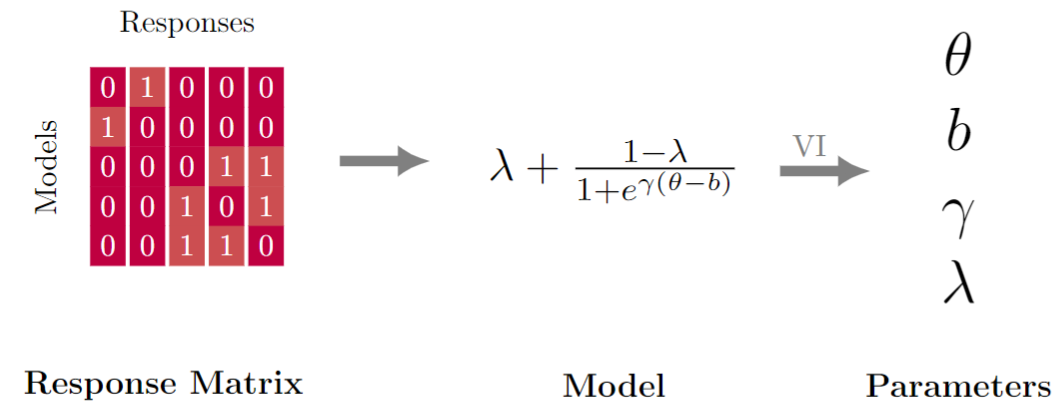}
    \vspace{-8pt}
    \caption{Overall Workflow}
    \vspace{-8pt}
    \label{fig:enter-label}
\end{figure}

\noindent \textbf{Reliability of IRT Parameter Estimates}
The reliability of the IRT parameter estimates can be verified by finding the Kendall correlations between the classical metrics and the IRT parameters. For instance, we can rank the models by both accuracy and ability and find the correlation between the two rankings. The same can be done with the difficulty ranking of images and the mean-item score. The correlations are shown in Table \ref{table:ranking}

    \begin{table}[h!]
    \footnotesize
    \centering
    \begin{tabular}{l|c|c}
    \toprule
    \textbf{IRT} & \textbf{Accuracy} & \textbf{Mean-Item}\\
    \textbf{Model} & \textbf{Ranking} & \textbf{Score Ranking}\\
    \midrule
    1PL & 0.99 & -0.96 \\
    2PL & 0.98 & -0.91 \\
    3PL & 0.98 & -0.9 \\
    \bottomrule
    \end{tabular}
    \caption{Correlation of IRT parameter estimates with classical metrics}
    \label{table:ranking}
    \end{table}

Another way to verify reliability is to use the IRT probability estimates to find the expected number of correct responses per model. For the 1PL model, the RMSE between the expected number of correct responses and the actual number is \textbf{158.71}

\begin{figure}[h!]
    \centering
    \begin{subfigure}
        \centering
        \includegraphics[width=0.45\textwidth]{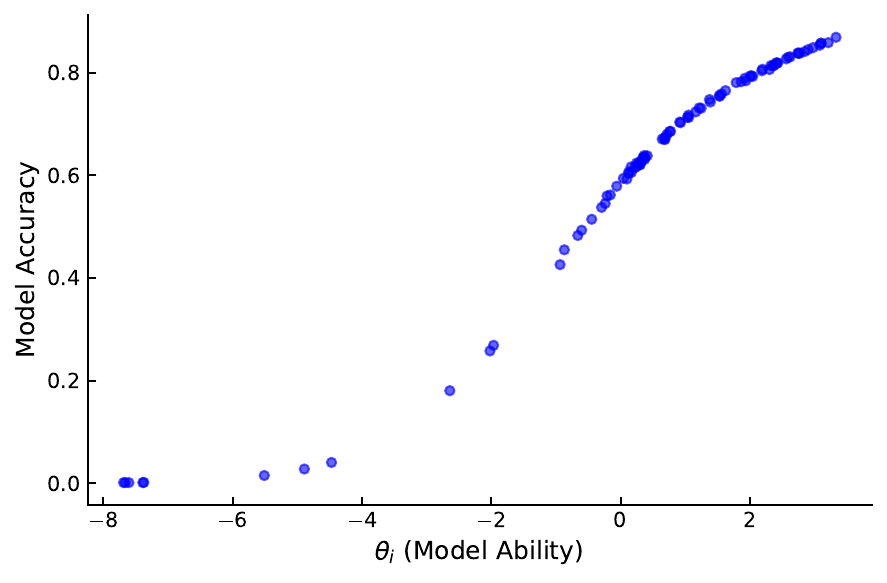}
        \caption{Scatter plot of model abilities and accuracies}
        \label{fig:1}
    \end{subfigure}
    \begin{subfigure}
        \centering
        \includegraphics[width=0.45\textwidth]{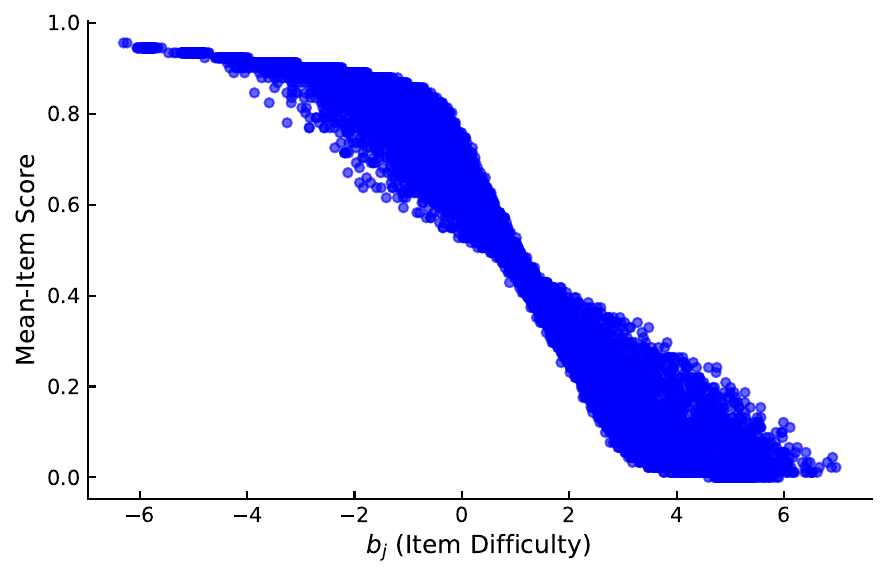}
        \caption{Scatter plot of image difficulties and scores}
        \label{fig:2}
    \end{subfigure}
    \label{fig:corr_scatterplots}
\end{figure}

The correlations in Table \ref{table:ranking} can be visualized through scatter plots, as depicted in Figs \ref{fig:1} and \ref{fig:2}. The plots exhibit a sigmoid shape, as expected, due to the form of the IRT equation.

It is important to distinguish between the IRT parameters and classical metrics like accuracy and mean item score ($P$). Generally, the classical metrics are highly correlated with the IRT parameters \citep{demars_item:10}, so ranking models by ability is equivalent to ranking them by accuracy. However, IRT parameters offer two key advantages: (i) The mathematical formulation of the IRT probability places ability ($\theta$) and difficulty ($b$) on the same scale, where the 50\% probability point corresponds to ability equaling difficulty. This reveals the relative distribution of models and items in a way accuracy cannot; (ii) IRT allows extracting other interpretable parameters like $\gamma$ and $\lambda$, offering further insights into the properties of items and the dataset.

\section{Experiments}
\label{sec_expts}
\noindent \textbf{Datasets.}
Our work focuses on classification, which has applications in many computer vision tasks. We perform most of our experiments on the validation set of ImageNet \citep{deng_imagenet:09}. Additionally, we conduct a subset of experiments on corrupted ImageNet, which includes 19 different corruptions at five severity levels (S5 being the most severe) \citep{hendrycks_benchmarking:19}. 

\noindent \textbf{Models.}
We use various models for the experiments, ranging from CNNs to transformer-based ones. We also use intermediate checkpoint models since simply using the strongest models leads to a weaker parameter fit \citep{martinez-plumed_IRT:19}. There are 91 models in total, including ConViT, ConvNeXt, DeiT3, DenseNet, EfficientNet, MaxViT, MobileNet, ResNet, RexNet, Swin Transformer, VGG, Xception, and ViTs. 57 models are trained locally using timm scripts \citep{rw_timm:19}.

\subsection{Assessing Model Calibration}
\label{sec: model_calibration}
Fundamentally, the IRT equation provides the probability that a model correctly classifies an image, which can be considered a "ground-truth" probability. Several studies \citep{northcutt_confidentlearning:21, klie_annerr:23} have shown that predicted class probabilities (softmax probabilities) effectively identify annotation errors. Building on this idea, we define a measure called overconfidence as follows: 
$$ \text{overconfidence}_{ij} = p^*(y_{ij} = 1) - \max_C p_i(l = C | j)$$
where $p^*(y_{ij} = 1)$ is the IRT probability, and $p_i(l = C | j)$ is the softmax probability predicted by model $i$ for class $C$, when image $j$ is input. For our experiment, we use the 2PL model, so $p^*(y_{ij} = 1)$ is given by Eq. \ref{eq:2pl}. Intuitively, the overconfidence measures the discrepancy between the model's estimate and the (potentially noisy) ground truth. 

For models with varying strengths, we plot the percentage of images with annotation errors and class overlap for given values of overconfidence. An image is defined to have class overlap if it shares its original label with additional classes. An image is considered to have an annotation error if its original label is incorrect. We use the Reassessed ImageNet labels for the true labels \citep{beyer_imagenet:20}.

\begin{figure}[t]
\centering
\begin{subfigure}{}
\centering
\includegraphics[width=0.4\textwidth]{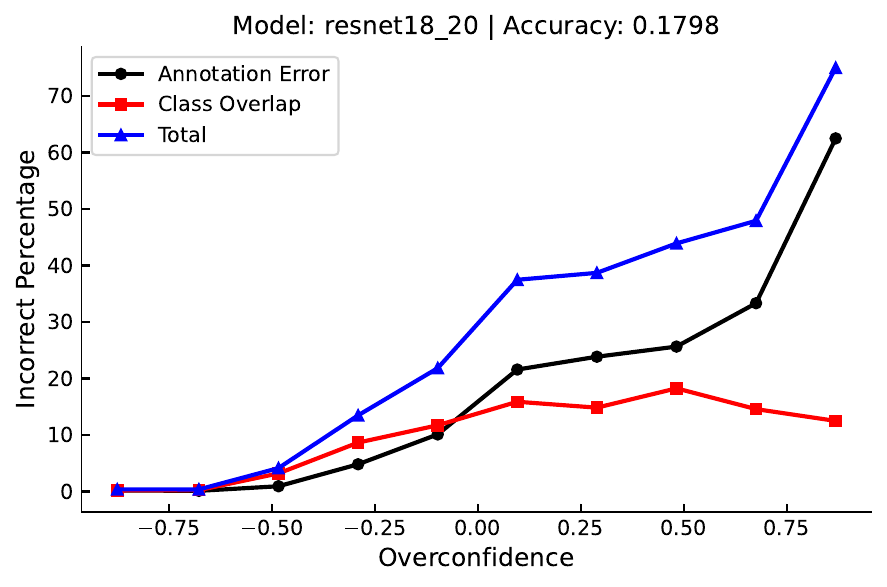}
\end{subfigure}%
\begin{subfigure}{}
\centering
\includegraphics[width=0.4\textwidth]{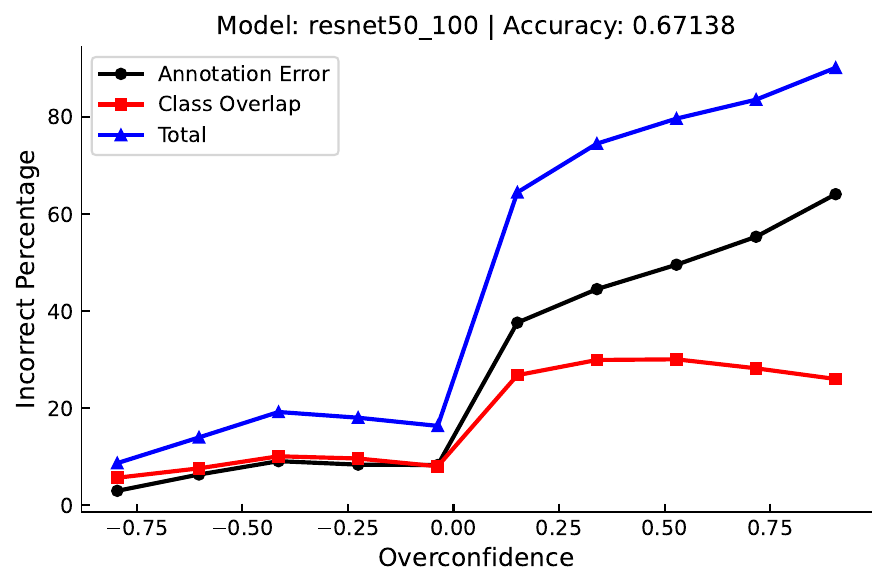}
\end{subfigure}%
\begin{subfigure}{}
\centering
\includegraphics[width=0.4\textwidth]{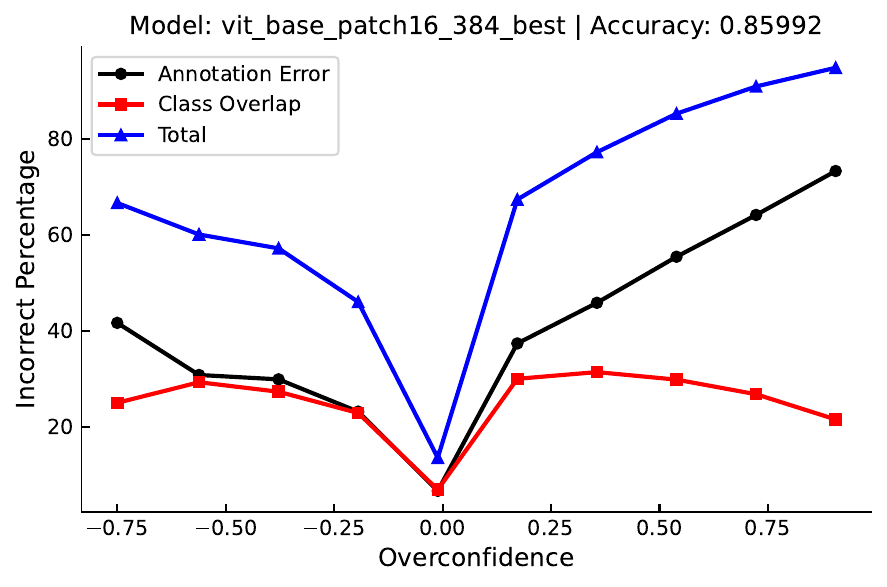}
\end{subfigure}%
\caption{Percentage of images with annotation errors and class overlap for given values of overconfidence across different models: \textit{(Top)} ResNet-18 (20 epochs) \textit{(Middle)} ResNet-50 (100 epochs) \textit{(Bottom)} ViT}
\label{fig:combined_error_percentages}
\end{figure}

The graphs in Fig. \ref{fig:combined_error_percentages} reveal a clear trend:
\vspace{-6pt}
\begin{itemize}[leftmargin=*]
\setlength\itemsep{-0.1em}
    \item Strong models are well-calibrated. When the overconfidence is 0, there are close to 0 label errors. However, even a slight deviation from this balance significantly increases the percentage of annotation errors. This sharp rise highlights the utility of overconfidence as an indicator of potential annotation errors. 
    \item For weaker models, the trend is less pronounced. While there is still an increase in annotation errors with increasing overconfidence, the relationship is not as steep or clear-cut. This could be due to poorer overall model calibration and less precise probability estimates.
\end{itemize}

In \ref{app:cont-irt-models}, we explore modelling the maximum softmax probabilities (confidences) using a continuous IRT model. We show inferring $b$ using both the confidences and response matrix results in difficulty values that are closer to the ground-truth.

\subsection{Dataset Complexity}
\label{sec: dataset_complexity}

\begin{figure}[t]
\centering
    \begin{subfigure}{}
    \centering
    \includegraphics[width=0.32\textwidth]{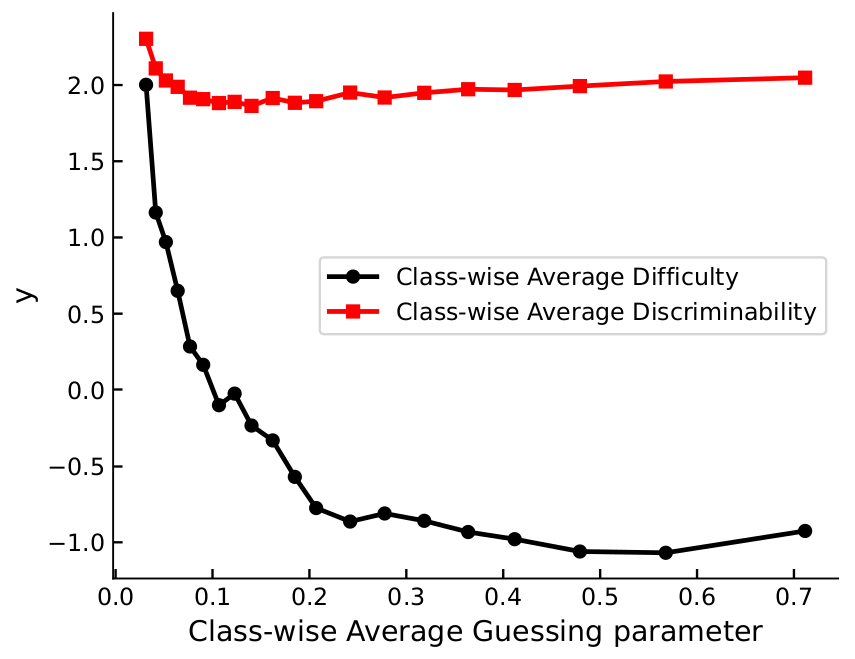}
    \end{subfigure}%

    \begin{subfigure}{}
    \centering
    \includegraphics[width=0.32\textwidth]{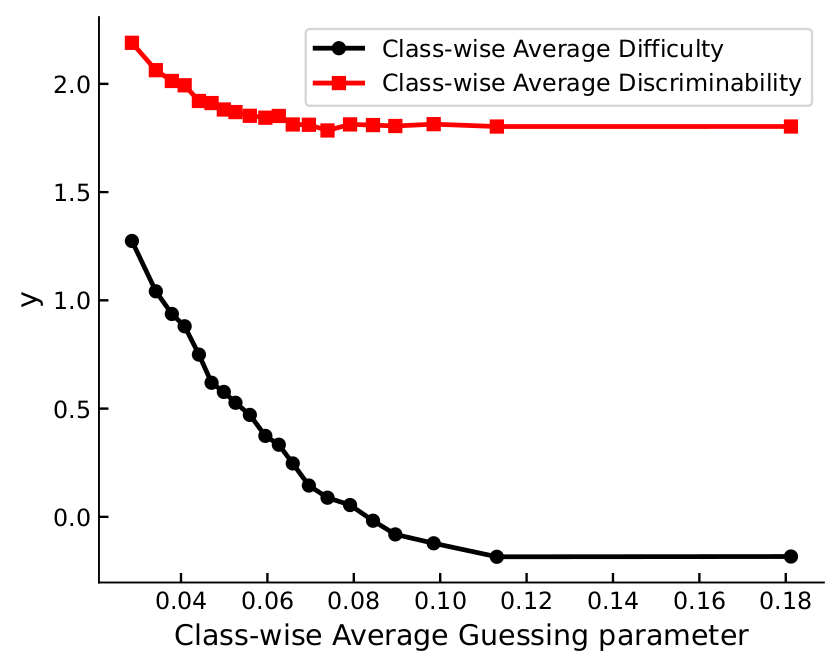}
    \end{subfigure}
\caption{Class-wise median guessing vs. class-wise median difficulty and discriminability of the Gaussian noise corruption: \textit{(Top)} Severity 1 \textit{(Bottom)} Severity 5}
\label{fig:gauss_noise}
\end{figure}

By definition, the guessing parameter measures the ease of guessing an item or image. In this section, we delve into the significance of the guessing parameter as a simple metric for assessing image dataset complexity. 

We focus on the median guessing parameter of each class of the ImageNet-C dataset. The idea behind using the median guessing parameter of each class is to measure the ease of guessing each class. As shown in Table \ref{table: guessing-parameter-table}, the median guessing parameter is inversely proportional to the severity level of the corruption, in other words, the complexity of the images. 

\begin{table}
\footnotesize
\begin{center}
\begin{tabular}{lccccr}
\toprule
ImageNet Class & S1 & S2 & S3 & S4 & S5 \\
\midrule
cardoon & 0.815 & 0.734 & 0.639 & 0.622 & 0.543 \\
pizza & 0.542 & 0.185 & 0.09 & 0.071 & 0.06 \\
jellyfish & 0.48 & 0.279 & 0.183 & 0.183 & 0.139 \\
leatherback\_turtle & 0.319 & 0.177 & 0.108 & 0.099 & 0.07 \\
Walker\_hound & 0.228 & 0.077 & 0.063 & 0.055 & 0.052 \\
assault\_rifle & 0.22 & 0.094 & 0.064 & 0.06 & 0.059 \\
\bottomrule
\end{tabular}
\caption{Median guessing parameters for 5 severity levels (S1 -$>$ S5) of frost corruption of 6 classes of ImageNet-C. }
\vspace{-15pt}
\label{table: guessing-parameter-table}
\end{center}
\end{table}

We also combine the median guessing parameter along with median difficulty and discriminability to study the effect of the guessing parameter on the other parameters. As shown in Fig \ref{fig:gauss_noise}, we observe that as the difficulty level rises, the guessing parameter decreases exponentially until reaching a plateau towards the end. This indicates that the images become difficult to guess as the complexity of the images increases. This also shows that the guessing parameter does not affect difficulty after a threshold. The discriminability almost has no effect due to the guessing parameter except the initial decrease.

Therefore, the study reveals that while the guessing parameter initially influences difficulty and discriminability, its impact diminishes significantly beyond a certain complexity threshold, suggesting limited interaction between these parameters at higher difficulty levels.

\subsection{Data Selection}
\label{sec: data_selection}
\begin{figure*}
    \centering
    \begin{minipage}{0.4\textwidth}
        \centering
        \includegraphics[width=\textwidth]{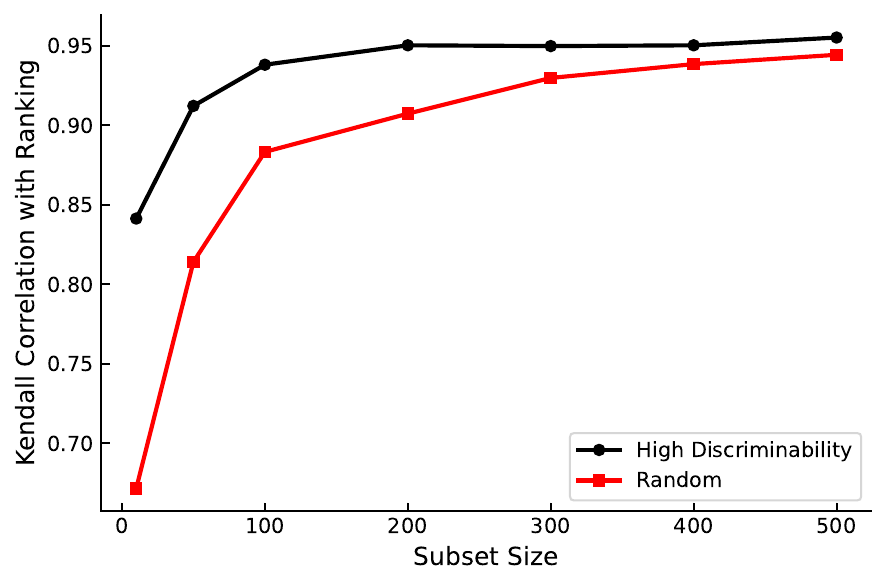}
        \caption{Correlation of rankings on small subset with overall rankings}
        \label{fig:correl}
    \end{minipage}\hfill
    \begin{minipage}{0.4\textwidth}
        \centering
        \includegraphics[width=\textwidth]{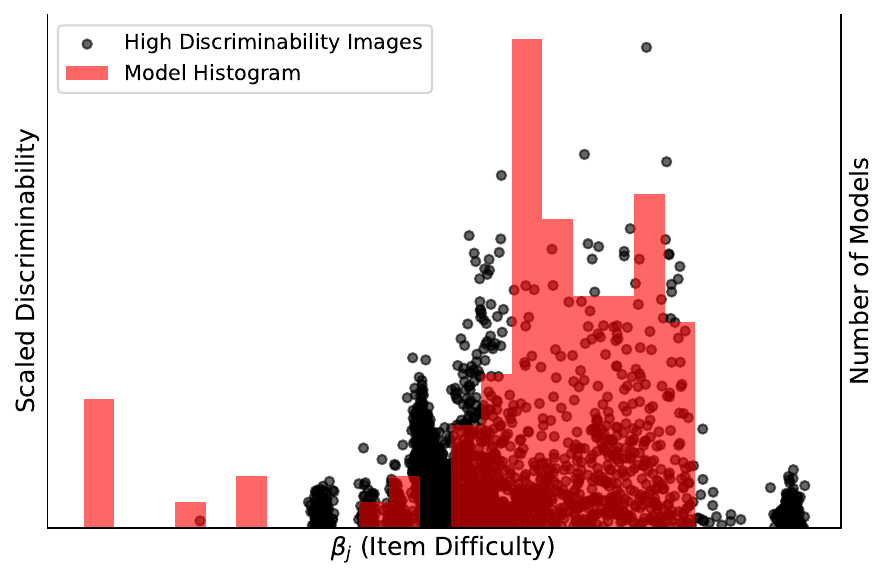}
        \caption{Histogram of model abilities overlaid on a scatter plot of highly discriminable images.}
        \label{fig:disc-hist}
    \end{minipage}
\end{figure*}

Item Response Theory (IRT) is a powerful tool for developing tests that effectively discriminate between examinees with varying abilities \citep{hambleton_item:85}. As previously discussed, a high discriminability parameter ($\gamma$) ensures that an item can reliably distinguish between models with abilities above or below the item's difficulty level. In the context of large-scale datasets for evaluating machine learning models, we can leverage this property to curate an extremely small yet highly informative subset of images. The curation of a highly discriminable set can be beneficial in situations where inference is expensive, to inexpensively compare a new model with a group of existing models. Figure \ref{fig:correl} illustrates the "informativeness" of such a selection. Even a subset comprising the 10 most discriminable images from the ImageNet validation set exhibits a remarkably high Kendall correlation of 0.85 with the overall model ranking obtained using the complete validation set. Furthermore, the IRT framework allows for fine-tuning the test subset to effectively discriminate between models within a specific ability range. This can be achieved by carefully selecting images whose difficulty levels ($b$) roughly match the abilities of the target group of models while also ensuring high discriminability ($\gamma$).

Interestingly, the most discriminable images have difficulty values that align well with the tested models. This is illustrated in Fig. \ref{fig:disc-hist}. Intuitively, when images are too easy or too difficult for a group of models, they can't be used to rank them. 

\section{Limitations and Research Directions}

We propose the following research directions and defer some results to the appendix while leaving others for future work:
\vspace{-6pt}
\begin{itemize}[leftmargin=*]
\setlength\itemsep{-0.1em}
    \item The assumption that ability is unidimensional is weak because different models can be better at different tasks. Multidimensional IRT breaks this assumption by treating abilities and difficulties as vectors; different models can excel at different traits. Preliminary results show that these models fit the data better.
    \item In the current formulation, IRT parameters require a response matrix for estimation. We can infer these quantities directly using a regressor or a neural network. There has been some work on this \citep{Martínez-Plumed_pred:22}, but the estimates are poor for more complex datasets like ImageNet.
    \item The abilities and difficulties derived from IRT can be leveraged for vote combination in model ensembles \citep{chen_item:19}. Our preliminary results, detailed in the appendix, demonstrate the potential of this approach.
\end{itemize}

\vspace{3pt}
\textbf{Reproducibility Statement}
\vspace{-6pt}
\begin{itemize}[leftmargin=*]
\setlength\itemsep{-0.1em}
    \item For the 1PL, 2PL, and 3PL models, the py-irt library \citep{lalor_pyirt:23} was utilized. A log-normal distribution was employed for discriminability instead of the standard normal distribution.
    \item The experimental IRT models were implemented using the Pyro probabilistic programming framework \citep{bingham_pyro:18}.
    \item  To promote reproducibility and enable further research in this area, the code will be made publicly available upon acceptance of this work.
\end{itemize}
\textbf{Broader Impact Statement.}
This work explores using Item Response Theory (IRT), a statistical framework traditionally employed in educational assessment, to provide nuanced insights into computer vision models and datasets. First, we introduce a novel approach to assessing model calibration and reveal that IRT can be used with model confidences to flag annotation errors. We then demonstrate how the guessing parameter can be utilized to evaluate dataset quality. Finally, we explore the discriminability parameter within the IRT framework and its application to data selection. Overall, this research makes strides toward assessing model calibration, gaining valuable insights into the difficulty and quality of datasets, and identifying the most informative data samples within these datasets. 





\bibliography{example_paper}
\bibliographystyle{icml2024}

\newpage
\appendix
\section{More about IRT}

\subsection{Variational Inference}
\label{app:vi}

We use variational inference to estimate the IRT parameters. We form the binary response matrix $Z^{n \times m}$ where $z_{ij} = 1$ implies that model $i$ classified image $j$ correctly, and $z_{ij} = 0$ indicates an incorrect response. We approximate the joint probabilities of the parameters $p(.)$ with a variational posterior $q_\phi(.)$. The variational distribution of the parameters is given in Eq. \ref{eq:mean_field_generative} below.

\begin{equation}
\label{eq:mean_field}
q_\phi (\theta, b, \gamma, \mu, \sigma) = q(\mu)q(\sigma) \prod_{ij} q(\theta_i)q(b_j)q(\gamma_j)  
\end{equation}
\begin{equation}
\label{eq:mean_field_generative}
\begin{aligned}
    \theta_i \sim \mathcal{N}(\mu_{\theta}, \tau_{\theta}^{-1}) \\ 
    b_j \sim \mathcal{N}(\mu_{b}, \tau_{b}^{-1}) \\
    \log{\gamma_j} \sim \mathcal{N}(\mu_{\gamma}, \tau_{\gamma}^{-1}) \\
    \lambda_j \sim U[0,1]
\end{aligned}
\end{equation}

As in \cite{natesan_bayesian:16}, we assumed that $\mu \sim \mathcal{N}$ and $\tau \sim \Gamma$. We assumed that the discriminability parameter $\gamma$ obeyed a log-normal distribution after observing a greater correlation of the fit parameters with the classical metrics. The parameters were fit by maximizing the evidence lower bound (ELBO)\footnote{https://github.com/nd-ball/py-irt}. 

An interesting point is that IRT models can be multidimensional, treating the abilities and difficulties as vectors.

\subsection{Multidimensional IRT Models}
The assumption that ability is unidimensional could be weak, as different students/models could be better at different traits. To accommodate this, the standard IRT equations can be extended \citep{de_theory:22}. The multidimensional 2PL model is given by Eq. \ref{eq:multidim-2pl}. Here, $v^i$ is the $i^{\text{th}}$ element of vector $\bm{v}$.

\begin{equation}
        \label{eq:multidim-2pl}
        p(y_{ij} = 1|\bm{\theta_i, b_j, \gamma_j}) = \frac{1}{1+ e^{-\sum_{i=1}^d\gamma_j^d(\theta_i^d - b_j^d)}}
\end{equation}


\subsection{Continuous IRT Models}
\label{app:cont-irt-models}
While modeling the probability of discrete responses in this paper, the framework of IRT can be extended to model the pdf underlying continuous responses \citep{noel_beta:07, chen_beta3irt:19}. We present here the model formulated in \cite{noel_beta:07}:
\begin{align*}
    \label{eq:beta-irt}
    f(y_{ij}|m_{ij}, n_{ij}) &= \beta(y_{ij}|m_{ij}, n_{ij}) \\
    &= \frac{\Gamma(m_{ij} + n_{ij})}{\Gamma(m_{ij})\Gamma(n_{ij})}y_{ij}^{m_{ij} - 1}(1 - y_{ij})^{n_{ij} - 1}
\end{align*}

where 
\begin{equation*}
    \label{eq:m-def}
    m_{ij} = \exp\left({\frac{\theta_i - b_j}{2}}\right)
\end{equation*}
\begin{equation*}
    \label{eq:n-def}
    n_{ij} = \exp\left(-{\frac{\theta_i - b_j}{2}}\right)
\end{equation*}
It can clearly be seen from the formula of the mean of the beta-distribution that:
\begin{align}
    \label{eq:exp_beta-irt}
    \mathbb{E}\left(Y_{ij}|m_{ij}, n_{ij}\right) &= \frac{m_{ij}}{m_{ij} + n_{ij}} \\
    &= \frac{\exp\left({\frac{\theta_i - b_j}{2}}\right)}{\exp\left({\frac{\theta_i - b_j}{2}}\right) + \exp\left(-{\frac{\theta_i - b_j}{2}}\right)} \\
    &= \frac{1}{1 + e^{-(\theta_i - b_j)}}
\end{align}

which is the same as Eq. \ref{eq:1pl}.

\begin{figure}[H]
    \centering
    \begin{subfigure}
        \centering
        \includegraphics[width=0.4\textwidth]{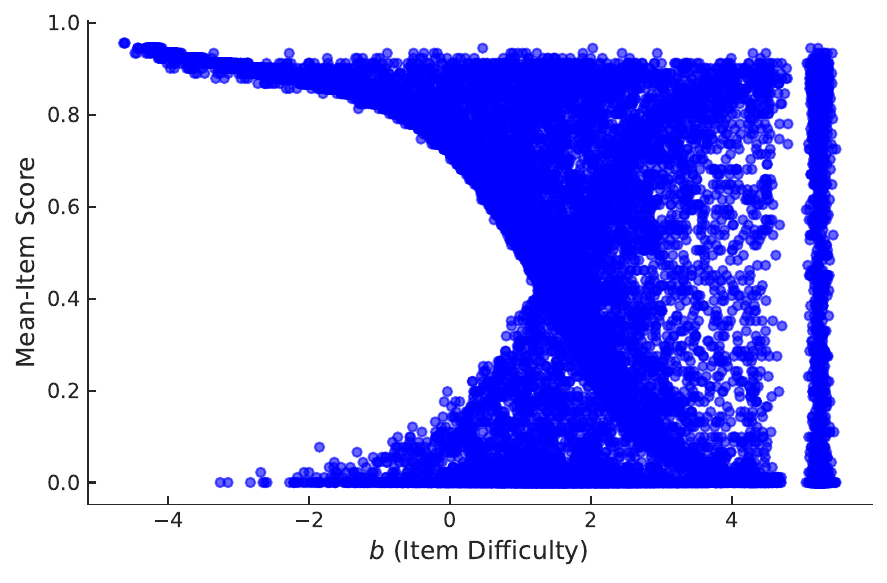}
        \caption{Scatter plot of difficulties and mean-item scores for the 2PL model}
        \label{fig:1-diff_mis_scatterplots}
    \end{subfigure}
    \begin{subfigure}
        \centering
        \includegraphics[width=0.4\textwidth]{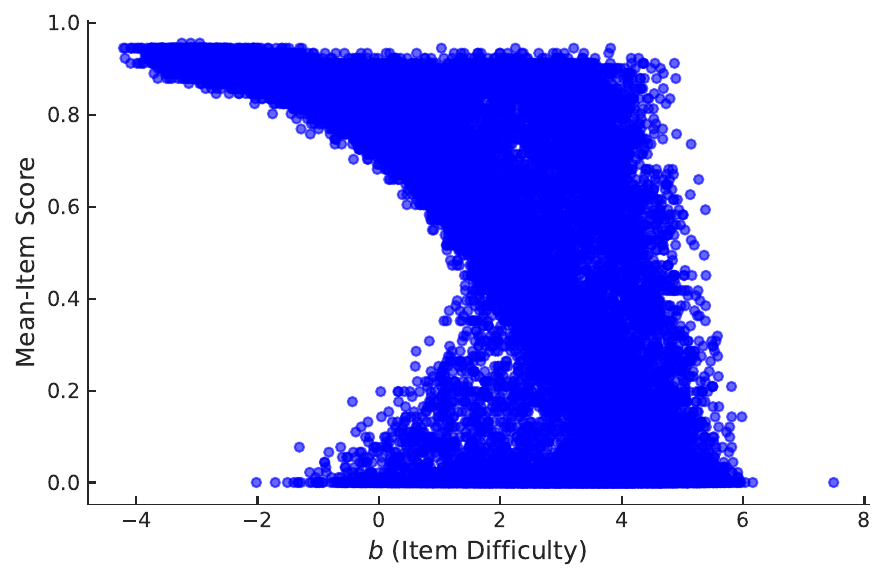}
        \caption{Scatter plot of difficulties and mean-item scores for the new model}
        \label{fig:2-diff_mis_scatterplots}
    \end{subfigure}
    \caption{Scatter plot of item difficulties vs. mean-item scores for the old and new IRT models}
    \label{fig:diff_mis_scatterplots}
\end{figure}

Here, we present an interesting experiment that incorporates this model. Observe that the IRT parameters obtained using the standard IRT models and the response matrix are solely a function of the response matrix. However, relying solely on the response matrix obscures information about an item's inherent difficulty or ease. For instance, all models might misclassify an image due to an annotation error, even if the image is relatively easy. This limitation is visualized in Fig. \ref{fig:1-diff_mis_scatterplots}, where the true Mean-Item Scores were obtained using ReaL labels \citep{beyer_imagenet:20}.

\begin{figure}
    \centering
    \includegraphics[width=0.45\textwidth]{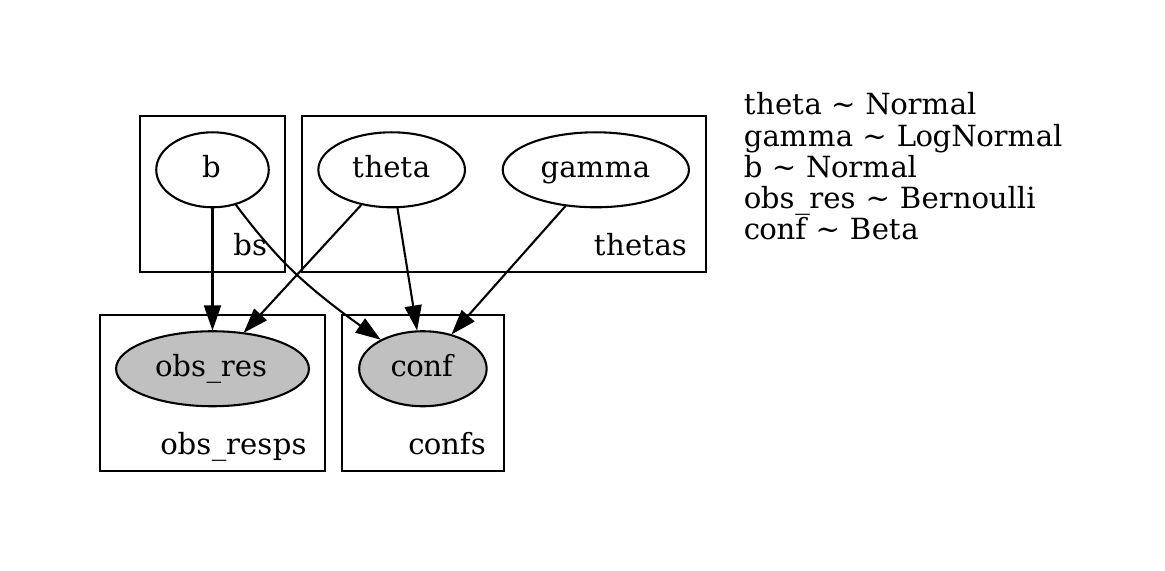}
    \caption{New IRT Model}
    \label{fig:new_irt_model}
\end{figure}

To address this limitation, we propose modifying the original IRT model by incorporating item difficulties to jointly predict both model confidences and responses. Intuitively, when a strong model exhibits high confidence for an item, it is likely an easier instance (\cite{northcutt_confidentlearning:21} use this concept to find label errors). Fig. \ref{fig:new_irt_model} demonstrates that by leveraging model confidences, this modified IRT model achieves a superior representation of the true item difficulties compared to the standard IRT formulation.

An interesting feature of this new IRT model is calibration. From \ref{fig:new_irt_model}, each vision model is associated with 1 $\gamma$ in addition to the $\theta$. Since the parameters are inferred using both \texttt{obs\_res} and \texttt{conf} (the maximum softmax confidences), $\gamma$ helps moderate the model confidences and bring them closer to the ground truth. We fit $\theta$ and $\gamma$ on a subset of 15000 images, freeze them, and then find the value of $b$ for the new images by fitting \textit{only} on the confidences.

\begin{figure}
    \centering
    \includegraphics[width=0.45\textwidth]{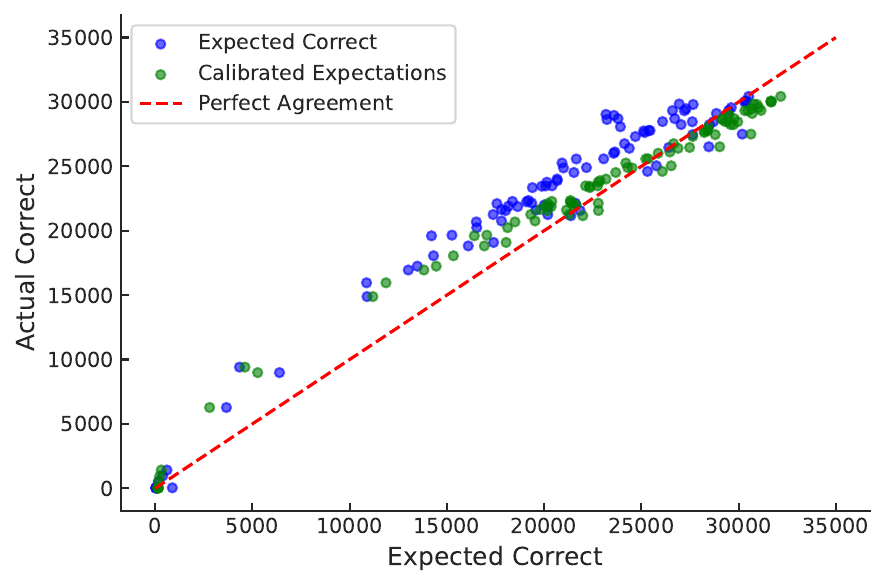}
    \caption{Expected Correct vs. Actual Correct}
    \label{fig:expected_vs_actual_correct}
\end{figure}

If we assume that $\mathbb{E}(\text{number of correct}) = \sum_i \max_C(\hat{p}(x_i))$, where $(\hat{p_C}(x_i))$ is the predicted probability for the $C^{\text{th}}$ class for image $x_i$, then using $\gamma$ to infer this probability helps calibrate the expected number of correct images, as visualized in \ref{fig:expected_vs_actual_correct}.

The original softmax probabilities give an ECE of 0.072, while the calibrated values give an ECE of 0.038.

\section{Experiments with Ensembles}

\begin{table}[h!]
    \footnotesize
    \centering
    \begin{tabular}{l|c|c}
    \toprule
    \textbf{Voting} & \textbf{Vanilla} & \textbf{ImageNet + }\\
    \textbf{Scheme} & \textbf{ImageNet} & \textbf{S5 Defocus-Blur}\\
    \midrule
    Majority Vote & 86.03 & 45.55 \\
    Strongest Model & 85.72 & 47.58 \\ 
    Softmax & 86.43 & \textbf{49.66} \\
    Regressor & \textbf{86.49} & 48.52 \\
    \bottomrule
    \end{tabular}
    \caption{Accuracies of different voting schemes}
    \label{table:ensemble-voting-scheme}
    \end{table}

Here, we utilize IRT parameters for weighted voting in ensembles \citep{chen_irtens:20, kandanaarachchi_unsupervised:21}. In particular, \cite{chen_irtens:20} propose a simple weighting scheme:

\begin{equation}
    w_i = \frac{e^{\theta_i}}{\sum_k e^{\theta_k}}
\end{equation}
Where the model abilities $\theta$ are obtained by inferring on a training set. We try this weighting scheme out on the ImageNet validation set; we infer using $15000$ randomly selected examples and find the accuracy on the remaining $35000$. We repeat the study on ImageNet with a severity 5 defocus-blur corruption. The results are reported in Table \ref{table:ensemble-voting-scheme}.

Inspired by \cite{Martínez-Plumed_pred:22}, we also explore using a regressor to infer parameters from the images and then using the parameters to form a weighted ensemble. By predicting the probabilities conditioned on the images, we can flexibly adjust the weights based on the image. A simple weighting scheme that implements this for model $i$ on image $j$ is $-\log(1 - p_{ij})$, where $p_{ij}$ is the probability that model $i$ gets image $j$ right.

\end{document}